\documentclass{article}

\usepackage[preprint, nonatbib]{neurips_2020}

\usepackage{amsmath,amsfonts,bm}

\def\eqref#1{equation~\ref{#1}}

\def\1{\bm{1}}

\DeclareMathAlphabet{\mathsfit}{\encodingdefault}{\sfdefault}{m}{sl}
\SetMathAlphabet{\mathsfit}{bold}{\encodingdefault}{\sfdefault}{bx}{n}

\newcommand{\R}{\mathbb{R}}

\usepackage[utf8]{inputenc} 
\usepackage[T1]{fontenc}    
\usepackage{hyperref}       
\usepackage{url}            
\usepackage{booktabs}       
\usepackage{amsfonts}       
\usepackage{nicefrac}       
\usepackage{microtype}      
\usepackage{mathrsfs}
\usepackage{multicol}
\usepackage{algorithm}
\usepackage{algorithmic}
\usepackage{adjustbox}
\usepackage{color}
\usepackage{mathrsfs}
\usepackage{amsmath} 
\usepackage[numbers]{natbib}
\usepackage{diagbox}
\usepackage{amssymb}
\usepackage[dvipsnames]{xcolor}
\usepackage{enumitem}
\newtheorem{thm}{Theorem}

\newtheorem{lemma}[thm]{Lemma}

\newtheorem{remark}{Remark}

\newtheorem{prop}{Proposition}
\newtheorem{problem}{Problem}
\newenvironment{proof}{Proof:}{\hfill$\square$}

\def\clip{\mathrm{clip}}
\def\BN{\mathrm{BN}}

\title{Optimal Quantization for Batch Normalization in Neural Network Deployments and Beyond}

\author{%
	Dachao Lin \\
	Center for Data Science \\
  	Peking University \\
  	Beijing, China \\
  	\texttt{lindachao@pku.edu.cn} \\
  	\And
  	Peiqin Sun \\
	Megvii \\  	
	Beijing, China \\
	\texttt{sunpeiqin@megvii.com} \\  	
  	\And
  	Guangzeng Xie \\
  	Center for Data Science \\
  	Peking University \\  	
	Beijing, China \\
	\texttt{smsxgz@pku.edu.cn} \\
	\And
  	Shuchang Zhou \\
	Megvii \\  	
	Beijing, China \\
	\texttt{zsc@megvii.com} \\
	\And	
  	Zhihua Zhang \\
  	School of Mathematical Sciences \\
  	Peking University \\
  	Beijing, China \\
  	\texttt{zhzhang@math.pku.edu.cn}
}

\begin{document}

\maketitle

\begin{abstract}
	Quantized Neural Networks (QNNs) use low bit-width fixed-point numbers for representing weight parameters and activations, and are often used in real-world applications due to their saving of computation resources and reproducibility of results.
	Batch Normalization (BN) poses a challenge for QNNs for requiring floating points in reciprocal operations, and previous QNNs either require computing BN at high precision or revise BN to some variants in heuristic ways. 
	In this work, we propose a novel method to quantize BN by converting an affine transformation of two floating points to a fixed-point operation with shared quantized scale, which is friendly for hardware acceleration and model deployment. 
	We confirm that our method maintains same outputs through rigorous theoretical analysis and numerical analysis. Accuracy and efficiency of our quantization method are verified by experiments at layer level on CIFAR and ImageNet datasets. 
	We also believe that our method is potentially useful in other problems involving quantization.
\end{abstract}

\section{Introduction}
Deep neural networks have achieved great success in many applications, such as computer vision \cite{krizhevsky2012imagenet}, speech recognition \cite{hinton2012deep} and Natural Language Processing \cite{bahdanau2014neural}. 
In computer vision, the most proposed architecture, Convolutional Neural Network (CNN), has demonstrated state-of-the-art results in many tasks.
However, CNN-based recognition systems need large amounts of memory and computational power, which may take up to weeks on a modern multi-GPU server for large datasets such as ImageNet \cite{deng2009imagenet}.
Hence, they are often unsuitable for smaller devices like embedded electronics, and there is a pressing demand for techniques to optimize models with reduced model size, faster inference and lower power consumption.

Accordingly, a variety of literature has made attention on the reduction of model size through the use of quantization \cite{lin2015neural, courbariaux2015binaryconnect, lin2016fixed}, low-rank matrix factorization \cite{denton2014exploiting, jaderberg2014speeding}, architecture pruning \cite{han2015deep, han2015learning}, etc. 
Quantization is one of the simpler ways to reduce complexity of any model with less precision requirements for weights and activations as well as speeding up the computation.

The methods for quantizing gradients, weights and activations \cite{lin2015neural, courbariaux2015binaryconnect, hubara2017quantized, zhou2016dorefa, rastegari2016xnor, banner2018scalable}, have achieved much closer performance to full precision networks. 
Whereas, after the previous efficient quantized neural network (QNN) training, there still has floating point operators in model inference particularly due to the float-type numbers in Batch Normalization (BN).

Since (deep) networks are hard trained, the BN operator \cite{ioffe2015batch} is usually used. Implementation of the conventional BN requires much computation of mean and variance, involving the sum of squares, square-root and reciprocal operations. These operators require float-type numbers for high precision. 
Previous attempts to use low precision networks do not use BN layers \cite{wu2018training} or keep them in full precision \cite{zhou2016dorefa}.
In addition, \citet{hubara2017quantized} proposed shift-based BN by replacing almost all multiplication with power-of-2 approximation and shift operations, and \citet{banner2018scalable} devised Range BN for the variance normalizing according to the range of input distribution. 
These modifications to BN for less computational costs are heuristic, confirmed only with experiments. 
Besides, modern hardware has less deployment cost with fixed-point style or perhaps some only support fixed-points \cite{ignatov2018ai}. Hence, previous QNN training methods with float-type BN layers fail to deploy on total fixed-number hardware.

In this paper we develop a direct way to handle the floating numbers in BN after obtaining a benchmark QNNs with previous training methods. 
We view conventional BN operator combined with several quantization operators used in feedforward step as an affine transformation with only two floating numbers.
In order to eliminate the floating numbers in BN, we start by considering the exact substitution of these two floating numbers to as less as possible fixed numbers. 
We recommend to quantize floating numbers with shared quantized scale and mathematically show that all floating points are able to convert to fixed-point operators in an idealized case.
In addition, we also demonstrate the lower bound and upper bound of the least possible quantized scale in our scheme.
After given the proposed quantized scale, we give few search attempts to decide remaining integers.    
By the way, our methods used for model deployment also guarantee the precision in QNNs' experiments, which can be seen as a supplement of modern QNN training.

Our main contribution is summarized as follows:
\begin{itemize}
	\item We propose a new method for quantizing BN in model deployment by converting the two floating points affine transformations to a fixed-point operation with shared quantized scale.
	\item We give theoretical guarantee of our quantization method, including the existence of quantized scale and the magnitude of optimal quantized bit-width. In addition, we accurately search the least possible quantized scale according to quantized bits to support our transformation numerically. 
	\item We conduct our method on CIFAR and ImageNet datasets based on benchmark quantized neural networks, showing little performance degradation if exists.
	\item Our scenario is restricted not only in BN, but other affine operators with floating numbers in similar quantization structure, such as several variants of BN: $L^p$ BN \cite{hoffer2018norm}, Instance Normalization \cite{ulyanov2016instance} and Range BN \cite{banner2018scalable}. 
\end{itemize}
 
The remainder of our paper is organized as follows. 
In Section \ref{pro-for}, we formulate the problem and present the motivation of our transformation.
In Section \ref{the-ana}, we give the equivalence of problem in Section \ref{pro-for} and show the upper bound and lower bound of satisfied solution through several properties. 
We then give a trivial search algorithm in Section \ref{exp-ana}, for finding accurate value of the solution and verifying previous results in Section \ref{the-ana}.
We also briefly discuss the precision loss in practical hardware implementation and validate numerically in common quantized neural networks in Section \ref{exp}.
Finally, we conclude our method in Section \ref{con}.
\section{Problem Formulation}\label{pro-for}
Quantization mainly focuses on compressing parameters in neural networks, such as weights in convolutional and fully-connected layers, activation outputs of hidden layers, and parameters in BN.   
A common design for QNN is uniform quantization \cite{zhou2017balanced, krishnamoorthi2018quantizing}, which makes quantized values evenly distributed among possible values, that is, 
\[ Q(x) = \frac{\lfloor \Delta x \rfloor}{\Delta}, x\in[x_{\min}, x_{\max}], \]
where $\Delta \in \mathbb{N}$ (the set of positive integers) is quantized scale, which measures the smallest gap between quantized values, and $\lfloor \cdot \rfloor$ means the floor function. The input $x\in \R$ is restricted to predefined interval $[x_{\min}, x_{\max}]$ due to limited computing power. 
Particularly, for $k$-bit uniform quantization, choose $\Delta=2^k-1$ for friendly hardware support and simply choose $x_{\min}=0, x_{\max}=1$ for non-symmetric quantizer,  $x_{\min}=-\frac{1}{2}, x_{\max}=\frac{1}{2}$ for symmetric quantizer.

The weights which should be uniformly quantized may first be mapped into the required input value range through normalized operator \cite{hubara2017quantized}
\[ x_i\rightarrow \frac{x_i}{2\max_{j}(|x_j|)}+\frac{1}{2}\in[0, 1], \]
or combined with hyperbolic tangent function \cite{zhou2016dorefa}
\[ x_i \rightarrow \frac{\tanh(x_i)}{2 \max_{j}(|\tanh(x_j)|)} + \frac{1}{2} \in [0, 1]. 
\]
Here the maximum is taken over all weights $x_j \in \R$ in the same layer. 

Activations are quantized more difficultly than weights due to nondifferentiable optimization incurred by quantizer functions \cite{cai2017deep, vanhoucke2011improving}. Several works \cite{lin2016fixed, rastegari2016xnor, zhou2016dorefa} have addressed this problem using bit-width allocation across layers or a continuous approximation to the quantizer function in the backpropagation step.

A common quantization approach for activations \cite{mishra2017wrpn, hubara2017quantized, zhou2016dorefa} quantizes output of the previous layer after clipping inputs into a limited interval, that is,
\[ Q_a(x, y_{\min}, y_{\max}) = Q(\clip(x, y_{\min}, y_{\max})),
\]
where $\clip(x,a,b) \triangleq \min\{\max\{x, a\}, b\}, \ b > a$, $y_{\min}$ and $y_{\max}$ are usually integers, and $Q(\cdot)$ is some quantizer function defined earlier.
For common used ReLU function, $y_{\min}=0$, $y_{\max}=1$ or $y_{\max}=6$ \cite{howard2017mobilenets, sandler2018mobilenetv2} (ReLU6) or some specific positive integers. 

Since weights and activations are of the form with constant quantization bits in denominator, feed-forward calculation is able to execute with complete fixed-point operation without regard of the fixed denominator, which achieves hardware acceleration.

Benchmark models employ BN for deep neural network training. The conventional BN operator is two affine transformations of four parameters:
\[ \BN(x, \mu, \sigma, \gamma, \beta) =  \gamma \frac{x-\mu}{\sigma} + \beta, 
\]
where $\mu$ and $\sigma$ are the mean and standard error estimators of output in the same layer and updated according to training batch statistics. 
Except conventional BN, several BN variants use similar transformation with different calculation method, such as $L^p$ BN \cite{hoffer2018norm} with $L^p$ norm of feature as $\sigma$, Range BN \cite{banner2018scalable} with maximum gap in feature as $\sigma$, Group Normalization \cite{Wu_2018_ECCV} and Instance Normalization \cite{ulyanov2016instance} with different part of feature for calculating $\mu$ and $\sigma$. Anyhow, all parameters are fixed during inference and deployment.

We emphasize that floating points come from the division
operation in BN and its variants, since $\mu$ is the running mean of batch feature in the same layer, $\beta$ and $\gamma$ are updated based on quantized gradients, but the reciprocal of $\sigma$ fails.

Previous state-of-the-art networks such as ResNet \cite{he2016deep} and VGG \cite{simonyan2014very} employ BN after convolutional layer or fully-connected layer, followed by an activation function.
Let $\mathbb{Z}$ denote the set of integer numbers. Suppose we use $\Delta = A$ and $W$ to quantize activations and weights.
Then from quantizer function $Q_a$, we can see outputs of previous layers are of the type $N_i / A$ with $N_i \in \mathbb{Z}$ and quantized weight $M_i/W$ with $M_i \in \mathbb{Z}$. 
Therefore, the outputs of convolutional or fully-connected layer are of the type $\sum_{i=1}^{r}\frac{M_i}{W}\cdot\frac{N_i}{A} + c$, where $r$ is the number related to the calculation of outputs in the next layer, and $c$ is the bias term if exists and may also be quantized. 
We record this output of the type $\frac{N}{AW}+c$ with $N\in \mathbb{Z}$. 
Here we do not identify specific corresponding position of parameters if no ambiguity.

After quantized convolutional or fully-connected layer is BN and quantized activation, then an element of the next layer outputs is  
\begin{equation}\label{two-affine}
	\begin{aligned}
	    & Q_a\left(\BN\left(\frac{N}{AW}+c, \mu, \sigma, \gamma, \beta \right), y_{\min}, y_{\max}\right) = \dfrac{\left\lfloor A \ \clip \left(\gamma \dfrac{\frac{N}{AW} + c-\mu}{\sigma} + \beta , y_{\min}, y_{\max}\right)\right \rfloor}{A}\\ 
		& = \dfrac{1}{A}
		\clip\left( \left\lfloor  \dfrac{N {+} AW \Big(\dfrac{\beta}{\gamma} \sigma + c - \mu \Big)}{ \dfrac{W}{\gamma}\sigma}
		\right \rfloor, Ay_{\min}, Ay_{\max} \right) \triangleq \dfrac{1}{A} \clip \left( \left\lfloor\dfrac{N+b}{t}\right \rfloor, Y_{\min}, Y_{\max} \right).
	\end{aligned}
\end{equation}

Here we exchange the order of $\clip$ and $\text{round}$ function because $y_{min}$ and $y_{max}$ are integers in second equality. We also simplify two-affine operator of BN to one-affine operator and replace some terms as follows: 
\[ Y_{\min} = Ay_{\min}, Y_{\max} = Ay_{\max}, \]
\[ b = AW(\dfrac{\beta}{\gamma} \sigma+c-\mu)\in \R, \; t = \dfrac{W}{\gamma}\sigma\in \R. \]

Generally, floating number operation will cause loss of precision, so this modification may lead to some error due to the limited computation precision. We will give experiments in subsequent section briefly. 

Without loss of generality, we assume $y_{\min}=0, y_{\max}=1$ clipped from ReLU function. Other cases can be obtained by shifting the clip range by
\[ \clip(x,a,b) = \clip(x-t,a-t,b-t)+t. \]
Then $Y_{\min}=0$ and $Y_{\max}=A$.

An empirical attempt is trying to make all operators only related to fixed-points, which means quantizing floating points $t$ and $b$ to some integers. 
A simple way is to let floating numbers $t$ and $b$ be integers respectively, which seems difficult to work because input variable $N \in \mathbb{Z}$ may vary in a wide range. 
A weaker requirement is to quantize $t$ and $b$ similar to quantizer function $Q$ and $Q_a$ do, namely approximate $t$ and $b$ by rational numbers:
\[ b\approx\frac{B_1}{B_2}, \quad  t\approx\frac{T_1}{T_2}. \]
Since this consideration would add two more numbers and no hardware support in advance if either $B_2$ or $T_2$ is decided. Another way is to consider whether the surrogates of $t$ and $b$ can share same quantized scale. For example, same prefixed denominator for whole potential floating numbers $t$ and $b$, which is consistent with previous scheme, because
\[ b\approx\frac{T_2B_1}{T_2B_2}\triangleq\frac{B}{K}, \quad t\approx\frac{T_1B_2}{T_2B_2}\triangleq\frac{T}{K}, \]
where we set $K = T_2B_2$, $B=T_2B_1$, $T=T_1B_2$. Then $t$ and $b$ share same pre-fixed quantized scale $K\in\mathbb{N}$.

Thus we want to replace the float numbers $t, b$ to integers $T, B, K$ for friendly hardware execution, where $K$ should not changed when $t$ and $b$ vary to support fixed hardware design.  

\begin{problem}\label{pro}
Given $A \in \mathbb{N}$, the problem is to find some $ K\in \mathbb{N}$ (as small as possible), such that for all $ t\neq 0, t, b \in \mathbb{R}$, there exist some $T\neq 0, T, B \in \mathbb{Z}$, such that for all $ N \in \mathbb{Z}$, 
\begin{equation}\label{ori-pro}
\mathrm{clip}\left(\left\lfloor \frac{N {+} b}{t} \right\rfloor, 0, A\right) = 
\mathrm{clip}\left(\left\lfloor \frac{N {\times} K {+} B}{T} \right\rfloor, 0, A\right).
\end{equation}
\end{problem}
In particular, we are able to examine whether $t, b$ can be directly converted to integers if $K=1$ already satisfies.

After having obtained $K$, we also need to decide the remaining quantized number $T, B$ based on the given $t, b$, which is the second problem we need to tackle.

\begin{problem}\label{pro2}
	Given $A \in \mathbb{N}$, $ t \neq 0, t, b \in \mathbb{R}$ and a suitable $ K\in \mathbb{N}$ which is one solution in Problem \ref{pro}, the problem is to find $T\neq0, T, B \in \mathbb{Z}$, such that for all $ N \in \mathbb{Z}$,
	\begin{equation*} \label{ori-pro2}
	\mathrm{clip}\left(\left\lfloor \frac{N {+} b}{t} \right\rfloor, 0, A\right) = 
	\mathrm{clip}\left(\left\lfloor \frac{N {\times} K {+} B}{T} \right\rfloor, 0, A\right).
	\end{equation*}
\end{problem}
Particularly, in the conventional $k$-bit quantization setting, we would set $A=2^k-1$ for some small $k\in \mathbb{N}$ and give the least bits number of satisfied $K$. 


\section{Theoretical Analysis}\label{the-ana}
In this section, we analyze Problem \ref{pro} and Problem \ref{pro2} with some propositions while leave detail in Appendix \ref{appendix}.

\begin{figure}[t]
	\centering
	\includegraphics[width=0.4\columnwidth]{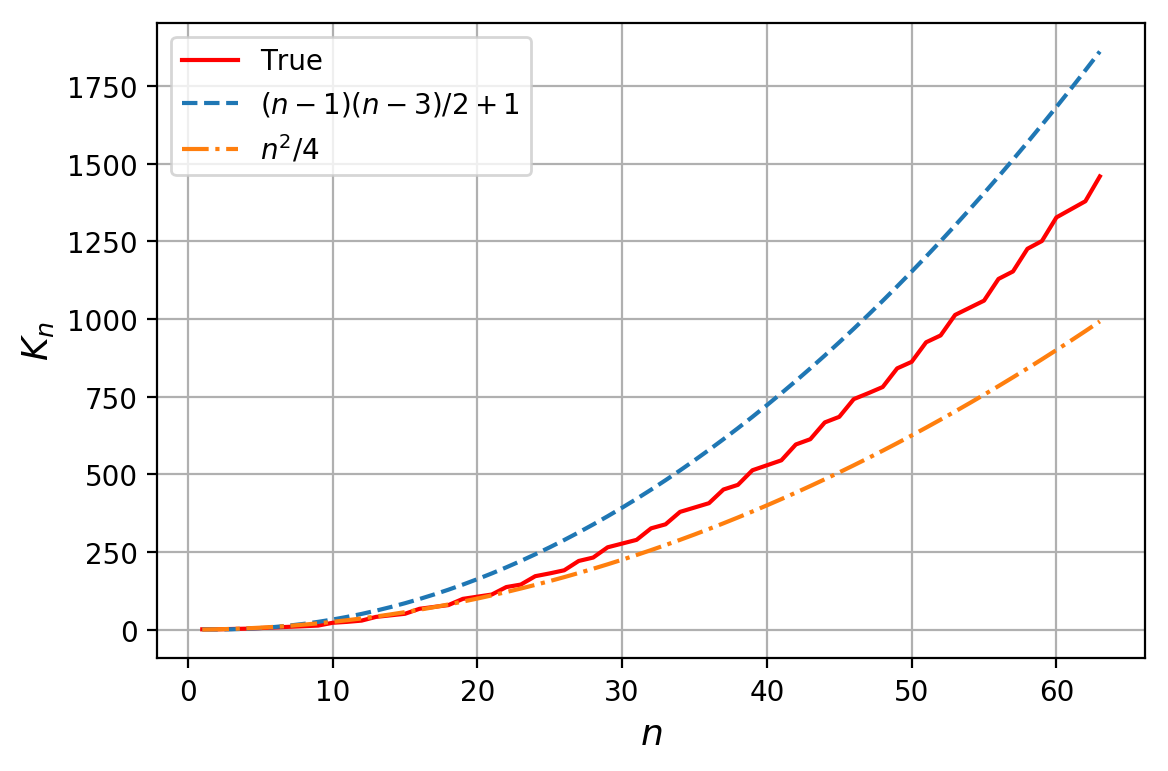}
	\hskip 0.1in
	\includegraphics[width=0.4\columnwidth]{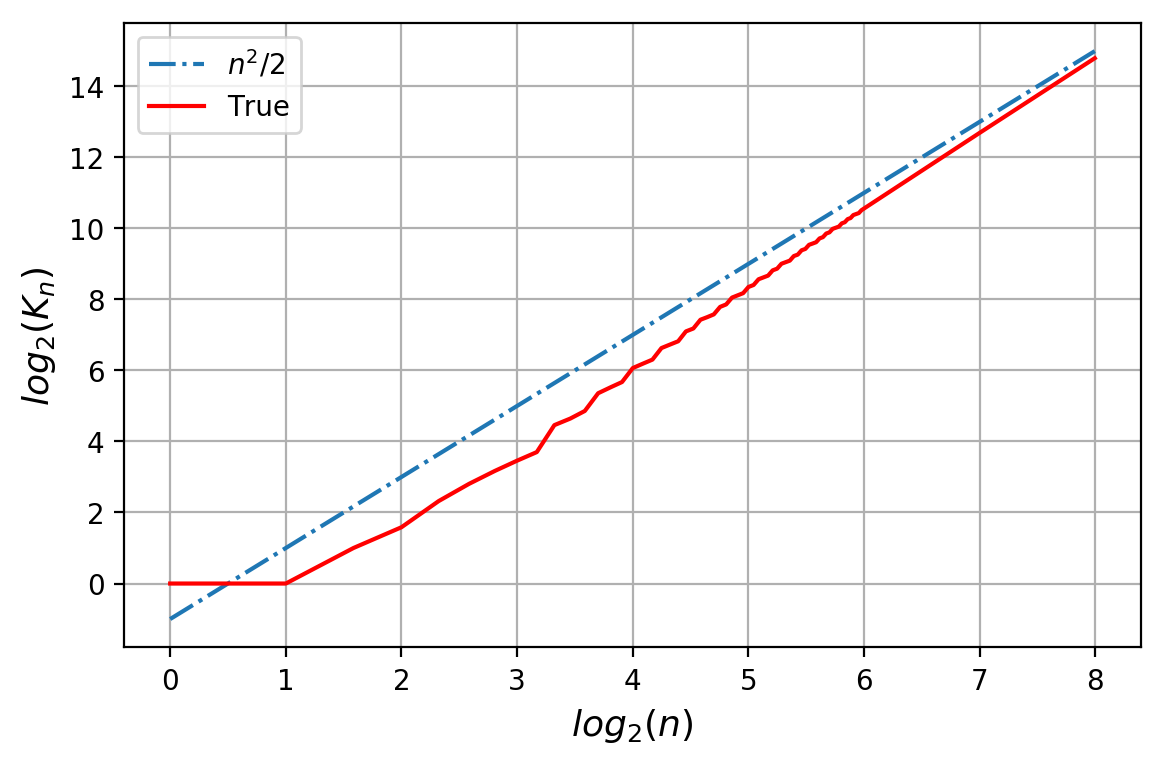}
	\vskip -0.1in
	\caption{Left: exact $K_n$ varies with $1\leq n \leq 63$ with upper bound and lower bound showed in Propositions \ref{pre-res} \& \ref{pre-res2}. Right: exact $K_n$ varies with $1\leq n \leq 255$ in log scale, showing the quadratic growth of $K_n$.}
	\label{fig:1}
\end{figure}

\subsection{Analysis of Problem \ref{pro}}
For Problem \ref{pro}, a direct way to deal with the randomness of $N$ is to specify all clipped intervals when $N$ varies. 

We conclude a simple version of Problem \ref{pro}  in Proposition \ref{2} (see Appendix \ref{app2} for proof).
\begin{prop}\label{2}
	Problem \ref{pro} is equivalent to finding such a $K \in \mathbb{N}$ that for all $ t \neq 0, t, b \in \R $, we can obtain $T \neq 0, T, B\in \mathbb{Z}$, s.t.
	\begin{equation}\label{p1}
		\lceil it-b \rceil = \Big\lceil \frac{iT-B}{K} \Big\rceil,\ \forall i \in \{1,2,\dots,A\}.
	\end{equation}
\end{prop}
Here $\lceil \cdot \rceil$ means the ceiling function.
We first analyze Eq. (\ref{p1}) without considering $T\neq0$ and leave the constraint in Appendix \ref{cons-T}, which have little influence in practice.

Set $ S_i = \lceil it-b \rceil $. According to Proposition \ref{2}, we only need to analyze the sequence $\{S_{i}\}_{i=1}^n$ (Here we replace $A$ by $n$ for notation simplicity). It seems hard to give analytic formula of $K$, though we only want to have a rough understanding of the least possible $K$ for future hardware support. Let the minimal $K \in \mathbb{N}$ which satisfies all sequences of length $n$ be $K_n$. First, we need roughly understand whether our transformation is reasonable.
The following proposition guarantees the existence of $K_n$ and also gives the magnitude of $K_n$ w.r.t.\ $n$.

\begin{prop}\label{pre-res}
	$ \forall n \in \mathbb{N} $, $K_n$ exists. Moreover, if $n > 4$, then as long as $K > \frac{(n-1)(n-3)}{2}$, we can find $T, B$ satisfying the requirements. Hence, $K_n \le \frac{(n-1)(n-3)}{2} + 1$.
\end{prop}

The key insight of the existence of $K_n$ is that while $K$ is large enough, then leave more choices to search for $T, B$ empirically on account of finite constraints in Eq. (\ref{p1}). The technique of proofs for the above proposition is to simply remove $T, B$ with conditions according to all possible sequences of $\{S_i\}_{i=1}^n$ and apply discreteness of integers. The proof is given in Appendix \ref{app2}.

As for lower bound of $K_n$, we can take some specific sequences to see how large the possible $K_n$ is. The insight of choosing several sequences $\{S_i\}_{i=1}^n$ is making the maximum gap between the items of such sequence small, in order to quick check the existence of $T$. 
\begin{prop}\label{pre-res2}
	If $n \ge 15$, we have $K_n \ge \frac{(n-1)^2}{4}$. Furthermore, if $n \ge 27$, then $K_n > \frac{n^2}{4}$. 
\end{prop}

\begin{remark}
	Propositions \ref{pre-res} and \ref{pre-res2} show that $K_n \asymp \Theta(n^2)$; more precisely, $\frac{1}{4}n^2 < K_n < \frac{1}{2}n^2$ except some small $n$. As for practical employment, when quantizing activations with $k$-bit, we need to choose $K$ at least $(2k-2)$-bit.
\end{remark}

\begin{remark}
	From Proposition \ref{pre-res}, we confirm that all $K > \frac{(n-1)(n-3)}{2}$ are the solution of Problem \ref{pro}. However, we will see that such a $K$ is not always sequential when 
	$K_n \leq K \leq \frac{(n-1)(n-3)}{2} $ in Appendix \ref{seq-k}.
\end{remark}

Besides, some special input range would obviously decrease the magnitude of $K_n$, such as the case Appendix \ref{sp-case} showed.

\subsection{Analysis of Problem \ref{pro2}}
Intuitively, given $K, b, t$ and leaving out floor function and clip function, we have 
\[ \frac{K}{T} \approx \frac{1}{t}, \quad \frac{B}{T} \approx \frac{b}{t}. \]
Hence, we can get $T\approx Kt, B\approx Kb$. Because of $T, B\in \mathbb{Z}$, we obtain $T \approx \lfloor Kt \rfloor$ (or $\lceil Kt \rceil$) and $B \approx \lfloor{Kb\rfloor}$ (or $\lceil Kb \rceil$). However, after subsequent search for all possible $T, B$, we find it is not always the nearest integer (see Proposition \ref{tbk2} and Appendix \ref{app8}). The gap of suitable $T, B$ and intuitive approximation depends on specific $t, b$. By the way, the intuitive way to obtain $T, B$ is correct most of the time.

To establish $T, B$  given $K$ that is suitable based on the previous understanding, we offer a conservative way to obtain all appropriate $T, B$ as Proposition \ref{tbk} and Algorithm \ref{getTB} (see Appendix \ref{algo}) mentioned .

\begin{prop}\label{tbk}
	Given $t, b \in \R$ and proper $K\in\mathbb{N}$, a pair of $T, B \in \mathbb{N}$ satisfy Problem \ref{pro2} if and only if there exists such a $T$  that the following conditions met:
	\begin{equation}
		\begin{aligned}
			\max_{i>j}\Big\lfloor\frac{S_i-S_j-1}{i-j}K \Big\rfloor &< T < \min_{i>j} \Big\lceil \frac{S_i-S_j+1}{i-j}K \Big\rceil, \\
			\max_{i} \left( iT- KS_i \right) &< \min_{j} \left( jT- KS_j\right) + K,
		\end{aligned}
	\end{equation}
	where $i, j \in \{1,2,\dots n\}$.
\end{prop}

Moreover, we show that the intuitive way to obtain $T, B$ is partially right in Proposition \ref{tbk2}, and give a loose bound of the possible ranges of $T$ and $B$ in Proposition \ref{tbk3} (see Appendix \ref{add-appendix}). In addition, for general quantizer range instead of $[0, A]$ in Problem \ref{pro}, we can apply similar analysis in Appendix \ref{gene-qua}  
\begin{table}[tb]
	\caption{Searched $K_n$ with commonly used number for $k$-bit quantization ($n=2^k-1$).}
	\label{ss_res}
	\centering
	\begin{tabular}{c|c|c|c|c|c|c|c}
		\toprule
		$ n $ & $K_n$ & $n$ & $K_n$ & $n$ & $K_n$  & $n$ & $K_n$ \\
		\midrule
		\bf{3} (2-bit)&2&\bf{7} (3-bit)&9&\bf{15} (4-bit)&51&
		\bf{31} (5-bit)&289\\
		\bf{63} (6-bit)&1459&\bf{127} (7-bit) & 6499&\bf{255} (8-bit)&28323 & \\
		\bottomrule
	\end{tabular}
\end{table}

\section{Numerical Computation}\label{exp-ana}

In the previous section we have discussed $K_n$ theoretically. In this section, we present how to compute the accurate $K_n$ to evaluate the theoretical analysis. 

For convenience, we are able to only consider $t \in [0, 1)$ and $\lceil t-b \rceil=1$ by Proposition \ref{sn-pro}. Then $S_1=1$, $0 \le S_{i+d}-S_i \le d$, $S_i\le i$, for $i \in \{1, \cdots, n-1\}, \ d \in \{1,\cdots, n-i\} $.

\begin{prop}\label{sn-pro}
Considering $t\in[0, 1)$ and $\lceil t-b \rceil=1$ is enough for finding $K_n$.
\end{prop}

For insurance purpose, we save all possible sequences $\{S_i\}_{i=1}^{n}$ as a naive idea, then to check every $K$ if the corresponding $T, B$ exist.
We generate the sequences of $T, B$ recursively based on the following result and the scope of $t, b$.

\begin{prop}\label{pro-seq}
	Suppose $\{S_i\}_{i=1}^{n+1}$ is a satisfied sequence of length $n+1$. Then $\{S_i\}_{i=1}^{n}$ is a satisfied sequence of length $n$. In addition, if $\{S_i\}_{i=1}^{n}$ is a satisfied sequence of length $n$, there must be a sequence of length $n+1$ and with the same previous $n$ terms.
\end{prop}

Now we turn to find $K_n$ in an accurate way. On account of Proposition \ref{pre-res2}, we start search from $\frac{1}{4}(n-1)^2$ when $n$ is larger than $15$. We use brute force and follow proposition below to search $K_n$ which satisfies all possible sequences $\{S_{i}\}_{i=1}^n$.

\begin{prop}\label{checktb}
	Given a sequence $ \{S_1,\dots,S_{n}\} $, if 
	\begin{equation}\label{tbpossible}
	\min_{j>i} \frac{j(S_i-1)-iS_j}{i-j} > \max_{j<i}\frac{j(S_i-1)-iS_j}{i-j},
	\end{equation}
	then there exist $t, b\in\R$ that can generate the  sequence above.
\end{prop}

The whole process is showed below: 

\begin{enumerate}
	\item First, produce candidate sequences. Note that from Proposition \ref{pro-seq}, we can get all the $\{S_{i}\}_{i=1}^n$ recursively.
	\item Second, record all satisfied sequences $\{S_{i}\}_{i=1}^n$. Given a candidate sequence $ \{S_1,\dots,S_{n}\} $, we use Proposition \ref{checktb} to confirm there exist $t, b$ which can reproduce this sequence. 
	\item Third, based on the obtained sequence $\{S_{i}\}_{i=1}^n$, check whether the given $K$ is satisfied through Proposition \ref{tbk}.
\end{enumerate}	

Due to the large amount of the possible sequences $\{S_{i}\}_{i=1}^n$, we suggest searching a small part of all sequences and running the window successively until all sequences are satisfied. The main pseudo-algorithm and the auxiliary pseudo-algorithms for search exact $K_n$ are all shown in Appendix \ref{algo}. 

\subsection{Experimental $K_n$ Results}\label{exp-ana2}

We give a short Table \ref{ss_res} corresponding to specific bit quantized activations. More $K_n$ with other $n$ please see Appendix \ref{all_kn}. 
Due to numerical errors, we also reexamine a wide range of input $N$ in $[-M-b, At-b+M]$ with large enough $M$ in order to ensure every possible input.

We also draw the growth trend of $K_n$ vs $n$ in Figures \ref{fig:1}.
Since saving all sequences is time-consuming and space-consuming, we only search $n$ in $[1, 63]$ and $\{127, \ 255\}$.
The magnitude depicted in the figures is consistent with Propositions \ref{pre-res} and \ref{pre-res2}, showing the quadratic growth of $K_n$.

In practice, there is no necessary requirement of finding $K_n$, since we only need some proper $K$ to convert the float-type parameters in BN as well as provide hardware friendly $K$, so not exactly $K_n$ is the best. 

\section{Experiments}\label{exp}
In this section we assess the effect of our BN quantization method through previous trained QNNs on ImageNet \cite{russakovsky2015imagenet} (we also try on CIFAR-10 \cite{krizhevsky2009learning} in Appendix \ref{app-cifar}). Our method is applied to model deployment at layer level, which is fitted for various quantized neural networks using the conventional BN or its variants. Besides, we aim at verifying accuracy variation through our modification for friendly hardware support, rather than obtaining high accuracy as previous quantization training methods do. 

There are two main concerns we should verify. First, the derivation in Eq. (\ref{two-affine}) absorbs the parameters of BN into the neighboring convolutional layer or fully-connected layer, which may have precision degradation.
We refer to $bt$ means absorbing BN parameters used in such layer though Eq. (\ref{two-affine}) based on quantized training models.
Second, because the original theoretical analysis is based on real numbers rather than floating numbers, we should check whether the theoretical results match with the expression of floating numbers. Though numerical experiments have already examined in Section \ref{exp-ana}, we display the final accuracy to support our method again. We adopt $BT$ which uses our fixed-points layerwise replacements in Problem \ref{pro} based on $bt$ in the following experimental tables. 

We use DoReFa-Net \cite{zhou2016dorefa}, one of efficient QNNs to train a quantized model for verification, and denote $QT$ as the accuracy with DoReFa-Net quantization methods. 

DoReFa-Net has low bitwidth weights and activations using low bitwidth parameter gradients to accelerate both in training and inference.
In order to get more suitable baselines, we only choose quantizing weights and activations while using full-precision gradients in training, and we do not quantize the final fully-connected layer for better training performance. 
We adopt $xWyA$ to represent $x$-bit quantized weights and $y$-bit quantized activations. 
We examine $2W4A$, $4W4A$ and $8W8A$ in subsequent experimental tables. 
We choose $K=64$ and $K=2^{16}$ for $4$-bit and $8$-bit quantized activations according to the discussion in Appendix \ref{seq-k}.

\subsection{Quantization for ImageNet Classification}
The next set of experiments study our method using VGG16 \cite{simonyan2014very} and ResNet18 \cite{he2016deep} on ImageNet dataset.
We adopt data augmentation including random cropping, horizontal flipping, lighting and color jittering with Adam optimizer using step-wise learning rate initialized from $0.3$ divided by $ 10 $ on each $ 30 $ epochs with total $ 100 $ epochs. After we obtain the trained QNN, we convert final model into  $bt$ and $BT$ mode we mentioned above. The main results are shown in Tables \ref{ex}.

\subsection{Accuracy and Complexity Analysis}
From Table \ref{ex}, using the QNN training method is able to get comparable performance when quantized even with $ 2W2A $ and $ 4W4A $. 
In addition, once a QNN has trained, we absorb the only floating-points in BN by our attempts. From the experimental results, $bt$ enjoys slight difference with the original quantized model $QT$, which means one affine transformation in Eq. (\ref{two-affine}) brings tiny disturbance. 
Moreover, we observe that our substitution in Problem \ref{pro} also introduces the same results between $bt$ and $BT$.
In principle, there should have diversity across $bt$ and $BT$ when encountering operators which excess the computer precision. We prefer to use our $BT$ case due to entire fixed-point implementation and if have, slight performance degradation.

Additionally, practical application of QNN mainly uses small bits (up to 8-bits to our knowledge with $n=255$ ) for quantization. From Proposition \ref{tbk} (or Proposition \ref{tbk3} in Appendix \ref{add-appendix}), a single conversion in worse case is $O(n)$, but time is able to be saved if we search around intuitive way $Kt$ and $Kb$, while we only need to convert model once with few BN parameters. 

\begin{table}[tb]
	\caption{Top-1 and Top-5 test accuracy of different quantized weight and activation bits with our transformation on ImageNet trained by VGG16 (top) and ResNet18 (bottom).} \label{ex}
	\centering
	\begin{tabular}{c|c|c|c|c|c|c}
		\toprule
		Bits  & \multicolumn{2}{c|}{2W4A} & \multicolumn{2}{c|}{4W4A} & \multicolumn{2}{c}{8W8A} \\		
		\midrule
		& Top-1 & Top-5 & Top-1 & Top-5 & Top-1 & Top-5 \\
		\midrule
		$ QT $   & 69.46 & 88.84 & 70.62 & 89.46 & 70.83 & 89.59 \\
		$ bt $   & 69.43 & 88.88 & 70.52 & 89.45 & 70.81 & 89.58 \\
		$ BT$    & 69.43 & 88.88 & 70.52 & 89.45 & 70.81 & 89.58 \\
		\midrule
		\multicolumn{7}{c}{VGG16} \\		
		\midrule
		\midrule
		$ QT $   & 65.94 & 86.54 & 68.15 & 88.09 & 68.67 & 88.18 \\
		$ bt $   & 65.99 & 86.55 & 68.15 & 88.11 & 68.66 & 88.17 \\
		$ BT$    & 65.99 & 86.55 & 68.15 & 88.11 & 68.66 & 88.17 \\
		\midrule
		\multicolumn{7}{c}{ResNet18}\\
		\bottomrule
	\end{tabular}
\end{table}
\section{Conclusion}\label{con}

In this paper we have investigated the problem of quantizing floating points in BN, combined with quantized convolutional or fully-connected weights and activations.
Noting that the conventional BN includes two affine transformations, we have made all floating points into one affine operator with only two floating-points. 
We have accordingly proposed to quantize each floating-point in the converted one-affine operator sharing a quantized scale.
We have shown that possible quantized scale is roughly twice of activation bits in theory and given numerical computation schemes of the corresponding substitution.  
Our approach enjoys the errorless performance in inference using high precision. 
The strategy we recommended displays efficient model deployment with complete fixed-point operators.

It is worth emphasizing that our quantization scheme is suitable for other affine operators which are also common in deep NNs. 
Beyond BN as well as the NN context, we  believe that our quantization scheme has potential applications in other problems that involve quantization. 


\section*{Broader Impact}
a) \& b) If the proposed quantization method is verified to be useful in numerous real applications by the engineers in the future, it will produce good impacts on model compression. Hence, previous style of QNNs would pay attention to entire fixed-point effective QNN design with BN. c) The method we proposed leverages all possible outputs and shows convincing results in theory, though our method may fails when the scope of fixed-point doesn't support the converted large range of $T, B$ that we seldom see this in practice. Besides, floating operation is likely to introduce precision error, which our method in first conversion step would encounter. d) Our method is data irrelevant.


\bibliographystyle{plainnat}
\bibliography{reference}

\appendix
\newpage
\appendix

\section{Additional Propositions} \label{add-appendix}
\begin{prop}\label{tbk2}
	Given $t, b \in \R$ and proper $K\in\mathbb{N}$, suppose $T, B$ which satisfy Eq. (\ref{ori-pro}) exist. Then either $T = \lfloor tK \rfloor$ or $ \lceil tK \rceil$ is candidate value, and either $B = \lfloor bK \rfloor$ or $ \lceil bK \rceil$ satisfies the requirements. However, the combined $T, B$ from these candidate values maybe not satisfy Problem \ref{pro2}.
\end{prop}

\begin{prop}\label{tbk3}
	Given $t, b \in \R$ and proper $K\in\mathbb{N}$, suppose $T, B$ which satisfy Eq. (\ref{ori-pro}) exist. Then the candidate values of $T$ and $B$ satisfy
	\[  Kt - \frac{2K}{n-1}< T < Kt + \frac{2K}{n-1}, \]
	\[  Kb-\frac{n+1}{n-1}K< B < Kb+\frac{n+1}{n-1}K. \]
\end{prop}

\section{Proof of Propositions} \label{appendix}
The following proofs may need use property of ceiling function below.
\begin{lemma}[The properties of ceiling function]\label{ceil-pro}
	\[ \lceil x\rceil + \lceil y\rceil\ge \lceil x+y\rceil\ge\lceil x\rceil+\lceil y\rceil-1.\]
	\[x \le \lceil x\rceil < x + 1.\]
\end{lemma}

\subsection{Proof of Proposition \ref{2}}\label{app1}
\begin{proof}
	Without loss of generality, we may assume $t > 0$, otherwise we can reverse $b,t,B,T$. Then we can see as $N \rightarrow \infty$, $\frac{K}{T}$ and $t$ must have the same sign, so $T \geq 0$ while we leave the case $T=0$ in Appendix \ref{cons-T}.
	
	Then according to the property of floor function, the original problem is equivalent to the follow situations: 
	\begin{equation}
	\left\{
	\begin{aligned}
	&\frac{N+b}{t} < 1           & \Leftrightarrow &   &         &\frac{NK+B}{T} < 1, \\
	i \le &\frac{N+b}{t} < i + 1 & \Leftrightarrow &   & \ i \le &\frac{NK+B}{T} < i+1, \ \forall i \in \{1,2,\dots,A-1\},\\
	&\frac{N+b}{t} \ge A     & \Leftrightarrow &   &         &\frac{NK+B}{T} \ge A.
	\end{aligned}
	\right.
	\end{equation}
	We can see $ \ \forall i \in \{1,2,\dots,A-1\}$,
	\[
	\left\{
	\begin{aligned}
	&N < t-b    & \Leftrightarrow &&& N < \frac{T-B}{K}, \\
	it-b \le    & N < (i + 1)t-b  & \Leftrightarrow && \ \frac{iT-B}{K} \le & N < \frac{(i+1)T-B}{K}, \\
	&N \ge At-b & \Leftrightarrow &&& N \ge \frac{AT-B}{K}.
	\end{aligned}
	\right.
	\]
	
	Hence $\lceil it-b \rceil = \lceil \frac{iT-B}{K} \rceil,\ \forall i \in \{1,2,\dots,A\}$. 
\end{proof}

\subsection{Proof of Proposition \ref{pre-res}}\label{app2}
\begin{proof}
	It is obvious that if $n = 1,2$, then $K_n = 1$.
	
	When $n > 2$, we only need to consider the problem in Proposition \ref{2}, that is, $\forall i\in \{1,2,\cdots,n\}$,
	\[
	\lceil it-b\rceil \ge \frac{iT-B}{K} > \lceil it-b  \rceil-1.
	\]
	Set $S_i=\lceil it-b\rceil$,
	\begin{equation} \label{o-eq}
	KS_i \ge iT-B > KS_i-K.
	\end{equation}
	Then \[ iT-KS_i \le B < iT-KS_i+K. \]
	Pay attention to $B \in \mathbb{Z}$, so that
	\[ \max_{i}(iT-KS_i) < \min_{i} (iT-KS_i+K), \]
	which means
	\[ (i{-}j)T-K(S_i{-}S_j)<K,\ \forall i,j\in \{1,2,\cdots,n\}. \]
	Analyzing the cases of $i > j$ and $i < j$, we obtain \\
	$\forall i,j\in \{1,2,\cdots,n\}, \ i>j$, 
	\begin{equation}\label{cond}
	\frac{S_i-S_j-1}{i-j}K < T < \frac{S_i-S_j+1}{i-j}K.
	\end{equation}
	It follows from Lemma \ref{ceil-pro} in Appendix \ref{appendix} that $\forall i, j \in \{1,\cdots,n\}$, $ i > j$,
	\begin{equation}\label{cond2}
	\frac{(S_i-1)-S_j}{i-j}< \frac{(it-b)-(jt-b)}{i-j}=t.
	\end{equation}
	$\forall l, m\in \{1,2,\cdots,n\}$, $l > m$,
	\[
	t = \frac{ (lt-b)-(mt-b)}{l-m} < \frac{S_l-(S_m-1)}{l-m}.
	\]
	Hence, 
	\[
	\frac{S_i-S_j-1}{i-j} < \frac{S_l-S_m+1}{l-m}.  \]
	Therefore $(S_l-S_m+1)(i-j)-(S_i-S_j-1)(l-m)\in \mathbb{N}$.\\
	We make
	\begin{equation}\label{eq_possible}
	\left(\frac{S_l-S_m+1}{l-m}-\frac{S_i-S_j-1}{i-j}\right)K > 1.
	\end{equation} 
	
	Then $T$ always exists because the scope of $T$ is larger than one and includes at least one integer.
	
	A naive idea is that when $ i - j \neq l - m $, 
	\begin{align*}
	& \frac{S_l-S_m+1}{l-m} -\frac{S_i-S_j-1}{i-j} \\ 
	& =\frac{(S_l {-} S_m + 1)(i{-}j)-(S_i {-} S_j {-} 1)(l{-}m)}{(l{-} m)(i{-}j)} \\
	& \ge  \frac{1}{(l-m)(i-j)} \ge\frac{1}{(n-1)(n-2)}.
	\end{align*}
	When $ i - j = l - m $ and $n > 2$, we have that
	\begin{align*}
	\frac{S_l-S_m+1}{l-m} -\frac{S_i-S_j-1}{i-j}&\ge\frac{1}{(l-m)}\ge\frac{1}{n-1}\\
	&\ge \frac{1}{(n-1)(n-2)}.
	\end{align*}
	So if $ K > (n-1)(n-2) $, we are able to find $T, B$ given any $t, b$. 
	Therefore, $K_n$ exists.
	
	\quad 
	
	Before the remaining proof, we need lemma below. 
	\begin{lemma}\label{lemma1}
		Suppose $ a,b,c,d \in \mathbb{N},a,c > 1 $ and $ab-cd = 1$, then
		$(b, c) = (a, d) = 1$ and
		\[ \forall n\in \mathbb{N}, n<a, \{\frac{dn}{a}\} \neq 0, \]
		\[ \forall n\in \mathbb{N}, n<c, \{\frac{bn}{c}\} \neq 0, \]
		where $\{ x \}$ is the fractional part of $x$.
	\end{lemma} 
	
	Now we turn to the left part of Proposition \ref{pre-res}.
	
	Using Proposition \ref{sn-pro}, we only need consider $t \in [0, 1)$.
	Let us take more precise analysis from Eq. (\ref{eq_possible}). Suppose
	\begin{equation}
	\Delta \triangleq \frac{S_l-S_m+1}{l-m}-\frac{S_i-S_j-1}{i-j} = \frac{(S_l-S_m+1)(i-j)-(S_i-S_j-1)(l-m)}{(l-m)(i-j)}.
	\end{equation}
	When $l-m = i-j$ and $n\geq5$,
	\[\frac{S_l-S_m+1}{l-m} -\frac{S_i-S_j-1}{i-j} \ge \frac{1}{l-m} \ge \frac{1}{n-1}\ge \frac{2}{(n-1)(n-3)}.\]
	So we only need to check $l-m \neq i-j$.
	\begin{enumerate}
		\item If $ (S_l-S_m+1)(i-j)-(S_i-S_j-1)(l-m)\ge3$, $n > 4$.
		\[\frac{S_l-S_m+1}{l-m} -\frac{S_i-S_j-1}{i-j} \ge \frac{3}{(l-m)(i-j)} \ge \frac{3}{(n-1)(n-2)}\ge \frac{2}{(n-1)(n-3)}.\]
		\item If $ (S_l-S_m+1)(i-j)-(S_i-S_j-1)(l-m) = 2 $, $n > 4$.
		
		Set $\alpha = S_l - S_m, \beta = S_i - S_j$.
		Use Proposition \ref{sn-pro}, then $0 \leq \alpha \le l - m, 0 \leq \beta \le i - j$.
		\begin{itemize}
			\item If $(l-m, i-j)=(n-1,n-2)$, then $(l, m) = (n, 1)$, \ $(i, j) = (n, 2)$ or $(n - 1, 1)$, then $\alpha \ge \beta$.
			
			Hence 
			\[2 = (\alpha+1)(n-2)-(\beta-1)(n-1) \ge 2n-3-\beta \ge 2n-3-(n-2) > 3. \] 
			No solution!
			\item If $(l-m, i-j)=(n-2,n-1)$, then $(i, j) = (n, 1)$, $(l, m) = (n, 2)$ or $(n - 1, 1)$, then $\alpha \ge \beta - 1$.
			
			Hence 
			\[2 = (\alpha + 1)(n-1)-(\beta-1)(n-2) \ge n - 2 + \beta > 2.\] 
			No solution!		
		\end{itemize}
		Therefore, $(l-m)(i-j) \le (n-1)(n-3)$
		\[\frac{S_l-S_m+1}{l-m} -\frac{S_i-S_j-1}{i-j} \ge \frac{2}{(l-m)(i-j)}\ge\frac{2}{(n-1)(n-3)}.\]
		
		\item If $ (S_l-S_m+1)(i-j)-(S_i-S_j-1)(l-m) = 1$, $n > 4$.
		
		Use Lemma \ref{ceil-pro},
		\[1 \ge (i-j)\lceil (l-m)t \rceil-(l-m)(\lceil (i-j)t \rceil-1) >(i-j)(l-m)t-(l-m)(i-j)t = 0.\]
		Hence 
		\begin{equation}
		(i-j)\lceil (l-m)t \rceil-(l-m)(\lceil (i-j)t \rceil-1)=1.
		\end{equation}
		
		Set $a = i-j, c = l-m$, which lead to
		\begin{equation} \label{mean-res}
		a\lceil ct \rceil-c(\lceil at \rceil-1)=1.
		\end{equation}
		So we can see
		\[\Delta \ge \frac{1}{(l-m)(i-j)}=\frac{1}{ca}.\]
		Next we will prove that $a, c$ satisfy the follow cases.
		
		Case1. If $a=1$ or $c=1$, then
		\[\Delta \ge \frac{1}{n-1}\ge \frac{2}{(n-1)(n-3)}.\]
		
		Case2. If $a + c \le n-1$, then
		\[\Delta \ge \frac{4}{(a+c)^2} \ge \frac{4}{(n-1)^2}\ge \frac{2}{(n-1)(n-3)}.\]
		
		From Case1, we may assume $a, c > 1$.
		Meanwhile take Eq. (\ref{mean-res}) into the condition $ (S_l-S_m+1)(i-j)-(S_i-S_j-1)(l-m) = 1$, we get $(a, c)=1$, and 
		\[ (S_l-S_m+1-\lceil (l-m)t \rceil)a = (S_i-S_j-\lceil (i-j)t \rceil)c.  \]
		
		Since $0\leq S_l-S_m+1-\lceil (l-m)t \rceil\leq1$, therefore
		\begin{equation}
		\begin{aligned}
		&\lceil lt-b \rceil-\lceil mt-b \rceil+1 =\lceil (l-m)t \rceil =\lceil c t \rceil, \\
		&\lceil it-b \rceil -\lceil jt-b \rceil =\lceil (i-j)t \rceil=\lceil a t \rceil.   
		\end{aligned}
		\end{equation}
		
		Hence 
		\begin{align}
		&\{mt-b\} +\{c t\} \le 1 \ and \ c t, mt - b \notin \mathbb{Z}.         \label{res1}\\      
		&\{jt-b\} +\{a t\} > 1 \ or  \ a t \in \mathbb{Z} \ or \ jt - b \in \mathbb{Z}.  \label{res2}
		\end{align}
		\begin{itemize}
			\item If $ at \in \mathbb{Z} $, then $\frac{a\lceil ct \rceil}{c} \ge \frac{act}{c} =at =\lceil at \rceil$. By Eq. (\ref{mean-res}),
			\[\frac{c(\lceil at \rceil-1)+1}{c}\ge \lceil at \rceil,\]
			so $c=1$, to Case1.
			\item If $ jt-b \in \mathbb{Z} $ or $\{jt-b\} +\{a t\} > 1$, by the previous fact that $at, ct \notin \mathbb{Z}$ and Eq. (\ref{mean-res}),
			we can get the range of $t$:
			\[\frac{\lceil at \rceil-1}{a} < t < \frac{\lceil ct \rceil}{c}.\] 
			We set 
			\[t = \frac{a\lceil ct\rceil-1+\delta}{ac} = \frac{c(\lceil at\rceil-1)+\delta}{ac},\delta \in (0,1),\]
			then 
			\[ \{ct\} = 1-\frac{1-\delta}{a} > \{at\} = \frac{\delta}{c}. \]
			The last inequality comes from $a, c > 1$.
			
			Reuse Eq. (\ref{res1}) and Eq. (\ref{res2}),
			\[ \{mt-b\} \le \frac{1-\delta}{a}, \{jt-b\} > 1-\frac{\delta}{c} \ or \ \{jt-b\} = 0. \]
			
			\begin{itemize}
				\item When $m = j$, because from Eq. (\ref{res1}), $ mt-b \notin Z$, so $\{jt-b\} \neq 0$, then
				\[ \frac{1-\delta}{a} \geq \{jt-b\}>1-\frac{\delta}{a}.\]
				Contradiction!
				\item When $m > j$, \[\{(m-j)t\}<\frac{1-\delta}{a}+\frac{\delta}{c}.\]
				If $m - j < a$,
				\[(m-j)\frac{\lceil at \rceil-1}{a}<(m-j)t<(m-j)\frac{\lceil ct \rceil}{c}.\]
				Then the range of $(m-j)t$ is $\frac{m-j}{ac}$ which less than $\frac{1}{c}$, and
				the fraction of the left point of $(m-j)t$ can't be zero by Lemma \ref{lemma1}. So $\frac{1}{a} \le \frac{1-\delta}{a}+\frac{\delta}{c}$, else $\frac{m-j}{ac} \geq \frac{1}{a}+\frac{1}{c} $, which means $ a + c \le n-1$, to Case2.\\
				Then $a \ge c$. As $(a,c)=1$, so $a > c$.
				\[\{(m-j)t\}<\frac{1-\delta}{a}+\frac{\delta}{c}<\frac{1}{c}.\]
				Due to $0<\frac{(m-j)\lceil ct \rceil}{c}-(m-j)t<\frac{1}{c}$, hence
				\[\left\{\frac{(m-j)\lceil ct \rceil}{c}\right\} = \frac{1}{c}.\] 
				Therefore $m-j = a - rc, \ r\in \mathbb{N}, rc<a$, but this time
				\[ \left\{ (m-j)t \right\}= \frac{\delta}{c}+\frac{(1-\delta)r}{a} \ge \frac{\delta}{c}+\frac{1-\delta}{a}.\] 
				No solution!\\
				Therefore $m - j \ge a$, so $c + a \le (l-m) +(m-j) = l-j\le n-1 $, to Case2.
				
				\item Similarly we can get same results when $m < j$ with
				\[\{(j-m)t\} > 1-\frac{1-\delta}{a} - \frac{\delta}{c}. \]
			\end{itemize}
		\end{itemize}
	\end{enumerate}	
	From previous discussion, we obtain
	\[\frac{S_l-S_m+1}{l-m} -\frac{S_i-S_j-1}{i-j} \ge\frac{2}{(n-1)(n-3)}.\]
	Hence, $T, B$ exists from Eq. (\ref{eq_possible}) when $n>4$ and $K>\frac{(n-1)(n-3)}{2}$.
\end{proof}

\subsection{Proof of Proposition \ref{pre-res2}}\label{app3}
\begin{proof}
	Consider $\{S_i\}_{i=1}^n $ through special choices of $t, b$ when $n > 8$.
	\begin{itemize}
		\item $ t=\frac{1}{n-2}$, $ b=\frac{3}{2n-4}-1 $, with sequence $\{S_i\}_{i=1}^n = \{1,2,2,\dots,2,2,2,3\}$,
		\item $ t=\frac{1}{n-3}$, $ b=\frac{3}{2n-6}-1 $, with sequence $\{S_i\}_{i=1}^n =\{1,2,2,\dots,2,2,3,3\}$, 
		\item $ t=\frac{1}{n-4}$, $ b=\frac{3}{2n-8}-1 $, with sequence $\{S_i\}_{i=1}^n = \{1,2,2,\dots,2,3,3,3\}$.
	\end{itemize}
	
	The corresponding satisfied $T$ is $T_1, T_2, T_3$. 
	From Eq. (\ref{cond}), the corresponding results are
	\[ \left(\frac{S_l-S_m+1}{l-m}\right)_{min} = \frac{1}{n-3},\frac{1}{n-4},\frac{1}{n-5}\ \forall l,m \in [n], \ l > m.\]
	\[ \left(\frac{S_i-S_j-1}{i-j}\right)_{max} = \frac{1}{n-1},\frac{1}{n-2},\frac{1}{n-3}\ \forall i,j \in [n], \ i > j.\]
	Therefore,
	\begin{equation}
	\left\{
	\begin{aligned}
	&\frac{K}{n-1}<T_1<\frac{K}{n-3},\\
	&\frac{K}{n-2}<T_2<\frac{K}{n-4},\\ 
	&\frac{K}{n-3}<T_3<\frac{K}{n-5}.\\
	\end{aligned}
	\right.
	\end{equation}
	Then
	\begin{align}
	&(n-3)T_1 < K < (n-1)T_1, \label{a1}\\
	&(n-4)T_2 < K < (n-2)T_2, \label{a2}\\ 
	&(n-5)T_3 < K < (n-3)T_3. \label{a3}
	\end{align} 
	From Eq. (\ref{a1}) and Eq. (\ref{a3}), then $(n-3)T_1 < (n-3)T_3$, so $T_3\ge T_1+1$. Take this into Eq. (\ref{a1}) and Eq. (\ref{a3}), then
	\[(n-1)T_1 \ge (n-5)T_3+2 \ge (n-5)(T_1+1)+2 \Rightarrow T_1\ge \Big\lceil\frac{n-3}{4}\Big\rceil.\]
	Then
	\[ K \ge (n-3)T_1 + 1\ge(n-3)\Big\lceil\frac{n-3}{4}\Big\rceil+1\ge\frac{(n-3)^2}{4}+1.\]
	If $ T_1= \big\lceil\frac{n-3}{4}\big\rceil, n > 9 $, from 
	\[(n-1)T_1 \ge (n-5)T_3+2 \Rightarrow T_3 = \Big\lceil\frac{n-3}{4}\Big\rceil+1.\]
	So either $T_1 \ge T_2$ or $T_2 \ge T_3$ can be satisfied.
	\begin{itemize}
		\item If $T_1 \ge T_2$, from Eq. (\ref{a1}) and Eq. (\ref{a2})
		\begin{align}
		&(n-3)T_1 < K < (n-2)T_1,\\
		&(n-5)T_3 < K < (n-3)T_3. 
		\end{align} 
		\[(n-5)T_3+2 \le (n-2)T_1 \Rightarrow (n-5)(\Big\lceil\frac{n-3}{4}\Big\rceil + 1 ) + 2 \le (n-2) \Big\lceil\frac{n-3}{4} \Big\rceil \Rightarrow n<15.\]
		\item If $T_2 \ge T_3$, from Eq. (\ref{a1}) and Eq. (\ref{a2})
		\begin{align}
		&(n-3)T_1 < K < (n-1)T_1,\\
		&(n-4)T_3 < K < (n-3)T_3. 
		\end{align} 
		\[(n-4)T_3+2 \le (n-1)T_1 \Rightarrow (n-4)(\Big\lceil\frac{n-3}{4}\Big\rceil+1)+2 \le (n-1)\Big\lceil\frac{n-3}{4}\Big\rceil \Rightarrow n<11.\]
	\end{itemize}
	So when $n \ge 15 $, then $ T_1= \big\lceil\frac{n-3}{4}\big\rceil+1$. Hence,
	\[ K_n \ge (n-3)T_1 + 1\ge (n-3)(\Big\lceil\frac{n-3}{4}\Big\rceil+1)+1\ge\frac{(n-1)^2}{4}.\]
	With the same idea further on the $T_1 = \big\lceil\frac{n-3}{4}\big\rceil+1$, we can get when $n \ge 27 $,\[ K_n \ge (n-3)T_1 + 1\ge (n-3)(\Big\lceil\frac{n-3}{4}\Big\rceil+2)+1 > \frac{n^2}{4}.\]
\end{proof}

\subsection{Proof of Proposition \ref{tbk}} \label{app4}
\begin{proof}
	From Eq. (\ref{o-eq}),
	we conclude that
	\[ \max_{i} \left( iT- KS_i \right) \leq B < \min_{j} \left( jT- KS_j\right) + K. \]
	Therefore $B$ exists when given $T$ if and only if 
	\[\max_{i} \left( iT- KS_i \right) < \min_{j} \left( jT- KS_j\right) + K. \]
	From Eq. (\ref{cond}) and $T \in \mathbb{N}$, we conclude that
	\[ \Big\lfloor\frac{S_i-S_j-1}{i-j}K \Big\rfloor < T < \Big\lceil \frac{S_i-S_j+1}{i-j}K \Big\rceil, \]
	where $i > j, i, j \in \{1,2,\dots n\}$. 
	Hence, $T$ exists if and only if
	\begin{equation}	\max_{i>j}\Big\lfloor\frac{S_i-S_j-1}{i-j}K \Big\rfloor < \min_{i>j} \Big\lceil \frac{S_i-S_j+1}{i-j}K \Big\rceil.
	\end{equation}
	In conclusion, $T, B$ exists if and ony if
	there exists such a $T\in\mathbb{N}$,
	\begin{equation}
	\begin{aligned}
	\max_{i>j}\Big\lfloor\frac{S_i-S_j-1}{i-j}K \Big\rfloor &< T < \min_{i>j} \Big\lceil \frac{S_i-S_j+1}{i-j}K \Big\rceil, \\
	\max_{i} \left( iT- KS_i \right) &< \min_{j} \left( jT- KS_j\right) + K,
	\end{aligned}
	\end{equation}
	where $i, j \in \{1,2,\dots n\}$.
\end{proof}

\subsection{Proof of Proposition \ref{sn-pro}}\label{app5}
\begin{proof}
	For the convenience of analysis, we can assume $t \in [0,1)$ and $ \lceil t - b \rceil =1 $ as long as we can minus an integer in both sides:
	\[
	\lceil i(t-\lfloor t \rfloor)-(b-b')\rceil = \left\lceil\frac{i(T-\lfloor t\rfloor K)-(B-b'K)}{K}\right\rceil , \  \forall i\in \{1,2,\cdots,n\},
	\]
	where $b'$ is the integer satisfies $ \left\lceil \{t\} - (b-b') \right\rceil =1 $.
	
	$S_1 = 1$, this is obvious by the assumption $0\le t < 1$ and $ \lceil t - b \rceil = 1 $.
	
	From Lemma \ref{ceil-pro}
	$S_{i+1} - S_i = 0 $ or $1$, so the sequence $\{S_{i}\}_{i=1}^n$ is non-decrease and $S_{i+d}-S_{i}\leq d, \ S_i\leq i$.
\end{proof}

\subsection{Proof of Proposition \ref{pro-seq}}\label{app6}
\begin{proof}
	First, as $\{S_i\}_{i=1}^{n+1}$ is a satisfied sequence with length $n+1$,then there exist $t,b$ satisfy $ S_i = \lceil it-b\rceil, \ \forall i \in \{1,2,\dots,n+1\} $, so $\{S_i\}_{i=1}^{n}$ is a satisfied sequence with length $n$ when we choose $t,b$ as the parameters.\\
	Second, when $\{S_i\}_{i=1}^{n}$ is a satisfied sequence with length $n+1$,then there exist $t,b$ satisfy $ S_i = \lceil it-b\rceil, \ \forall i \in \{1,2,\dots,n\} $, so $\{S_1,S_2,\dots,S_n,\lceil (n+1)t-b\rceil\}$ is a satisfied sequence with length $n+1$ and the same previous $n$ terms.
\end{proof}

\subsection{Proof of Proposition \ref{checktb}}\label{app7}
Note that  
\[ S_i-1<it-b\le S_i,\  \forall i \in \{1,2,\dots,n\}. \]
Then 
\[\frac{S_i-1+b}{i} < t \le \frac{S_i+b}{i} ,\ \forall i \in \{1,2,\dots,n\}.\]
So $t\in \R$ exists as long as
\[\frac{S_i-1+b}{i} < \frac{S_j+b}{j} ,\ \forall i,j \in \{1,2,\dots,n\},\]
which means that
\[
\left\{
\begin{aligned}
& b < \frac{j(S_i-1)-iS_j}{i-j}, \ j > i,\\
& b > \frac{j(S_i-1)-iS_j}{i-j}, \ j < i.
\end{aligned}
\right.
\]
Therefore, $b$ exists if and only if 
\begin{equation}
\min_{j>i} \frac{j(S_i-1)-iS_j}{i-j} > \max_{j<i}\frac{j(S_i-1)-iS_j}{i-j}.
\end{equation}

\subsection{Proof of Proposition \ref{tbk2}} \label{app8}
\begin{proof}
	Once we know $T,B$ exist, then from Eq. (\ref{cond2}), and Lemma \ref{ceil-pro}, $\forall i,j,l,m\in [n], i>j,l>m$,
	\begin{equation}\label{t-eq}
	\frac{S_i-S_j-1}{i-j}K < tK < \frac{S_l-S_m+1}{l-m}K.
	\end{equation}
	Since there exist $T$ satisfy requirements, so there exists an integer in the interval 
	\[ \bigg(\max_{i>j}\frac{S_i-S_j-1}{i-j}K,\  \min_{i>j}\frac{S_i-S_j+1}{i-j}K \bigg).\]
	From Eq. (\ref{t-eq}), $tK$ is in the interval, so the neighboring integer $\lfloor tK \rfloor$ or $\lceil tK \rceil$ must have at least one satisfies Eq. (\ref{ori-pro}). 
	
	Similarly, from Eq. (\ref{o-eq})
	\[ \frac{KS_i-K+B}{i} < T \le \frac{KS_i+B}{i}, \ \forall i\in \{1,2,\cdots,n\}.  \]
	\[ \frac{KS_i-K+B}{i} < \frac{KS_j+B}{j}, \ \forall i, j\in \{1,2,\cdots,n\}.  \]
	\[ \left(\frac{1}{i}-\frac{1}{j}\right)B < K\left(\frac{S_j}{j}-\frac{S_i-1}{i}\right), \ \forall i, j\in \{1,2,\cdots,n\}.  \]
	\[ K \left(\frac{S_i-1}{i}-\frac{S_j}{j}\right) \bigg/ \left(\frac{1}{j}-\frac{1}{i}\right)   < B < K \left(\frac{S_l}{l}-\frac{S_m-1}{m}\right) \bigg/ \left(\frac{1}{m}-\frac{1}{l}\right), \ \forall i > j, l > m. \]
	Since 
	\[  \frac{S_l}{l}-\frac{S_m-1}{m}>\frac{lt-b}{l}-\frac{mt-b}{m} =\left(\frac{1}{m}-\frac{1}{l}\right)b,\]
	\[ \frac{S_i-1}{i}-\frac{S_j}{j} < \frac{it-b}{i}-\frac{jt-b}{j} =\left(\frac{1}{j}-\frac{1}{i}\right)b. \]
	Hence
	\[K \left(\frac{S_i-1}{i}-\frac{S_j}{j}\right) \bigg/ \left(\frac{1}{j}-\frac{1}{i}\right)   < bK < K \left(\frac{S_l}{l}-\frac{S_m-1}{m}\right) \bigg/ \left(\frac{1}{m}-\frac{1}{l}\right), \ \forall i > j, l > m.\]
	Since there exist $B$ satisfy requirements, so that there exists an integer in the interval 
	\begin{equation}\label{b-eq}
	\bigg(\max_{i>j}K \left(\frac{S_i-1}{i}-\frac{S_j}{j}\right) \bigg/ \left(\frac{1}{j}-\frac{1}{i}\right) ,\  \min_{i>j}K \left(\frac{S_i}{i}-\frac{S_j-1}{j}\right) \bigg/ \left(\frac{1}{j}-\frac{1}{i}\right)\bigg).
	\end{equation}
	From Eq. (\ref{b-eq}), $bK$ is in the interval, so the neighboring integer $\lfloor bK \rfloor$ or $ \lceil bK \rceil$ must have at least one satisfies Eq. (\ref{ori-pro}). 
	
	However, the combined pair $T\in \{\lfloor bK \rfloor, \ \lceil bK \rceil\} $, and  $ B\in \{ \lfloor tK \rfloor, \ \lceil tK \rceil\} $ may don't satisfy Problem \ref{pro2}. \\
	For example, take $b = 0.198, t = 0.618, n = 15, K = 64$, all possible sequences $(T, B) = (39, 6), (39, 7), (39, 8)$, while $\lfloor tK \rfloor = 39, \lfloor bK \rfloor = 12$.
\end{proof}

\subsection{Proof of Proposition \ref{tbk3}} \label{app9}
\begin{proof}
	From Eq. (\ref{cond2}) and Lemma \ref{ceil-pro},
	\[ \frac{S_l-S_m+1}{l-m} < \frac{(lt-b+1)-(mt-b)+1}{l-m}=t + \frac{2}{l-m}. \]
	Therefore
	\[ \min_{l>m}\frac{S_l-S_m+1}{l-m} < t + \min_{l>m}\frac{2}{l-m} = t + \frac{2}{n-1}.\]	
	Similarly,
	\[\max_{i>j} \frac{S_i-S_j-1}{i-j} > \max_{i>j} \frac{(it-b)-(jt-b+1)-1}{i-j} = t + \max_{i>j}-\frac{2}{i-j} = t-\frac{2}{n-1}. \]
	Hence
	\[  Kt - \frac{2K}{n-1}< T < Kt + \frac{2K}{n-1}.\]
	From Eq. (\ref{b-eq})
	\[ \left(\frac{S_i-1}{i}-\frac{S_j}{j}\right) > \frac{it-b-1}{i}-\frac{jt-b+1}{j}=-\left(\frac{1}{i}-\frac{1}{j}\right)b-\left(\frac{1}{i}+\frac{1}{j}\right). \]
	\[\max_{i>j} \left(\frac{S_i-1}{i}-\frac{S_j}{j}\right) \bigg/ \left(\frac{1}{j}-\frac{1}{i}\right) > \max_{i>j}b-\left(\frac{i}{j}+1\right) / \left(\frac{i}{j}-1\right) = b-\frac{n+1}{n-1}. \]
	\[\min_{i>j} \left(\frac{S_i}{i}-\frac{S_j-1}{j}\right) \bigg/ \left(\frac{1}{j}-\frac{1}{i}\right) > \min_{i>j}b + \left(\frac{i}{j}+1\right) / \left(\frac{i}{j}-1\right) = b+\frac{n+1}{n-1}. \]
	Hence
	\[ Kb-\frac{n+1}{n-1}K<B<Kb+\frac{n+1}{n-1}K.\]
\end{proof}

\section{Searched $ K_n $}\label{all_kn}
In this section, we list more $K_n$ with various $n$ in Table \ref{se_res}. Since searching accurate $K_n$ is time-consuming, we only find large bit quantized scale ($n=2^k{-}1$) when $n$ is large.

\begin{table}[t]
	\caption{Searched $K_n$ with various $n$. Some useful numbers for $k$-bit quantization ($n=2^k-1$) are displayed in bold.} \label{se_res}
	\vskip 0.15in
	\centering
	\begin{tabular}{cc|cc|cc|cc|cc}
		\toprule
		$ n $ &$K_n$& $ n $ &$K_n$& $ n $ &$K_n$& $ n $&$K_n$ &$ n $& $K_n$ \\
		\midrule
		1     &  1  &  14   &  46 &   27  & 221 & 40 & 529 &  53    & 1013 \\
		2     &  1  &\bf{15}&  51 &   28  & 232 & 41 & 545 &  54    & 1036 \\
		\bf{3}&  2  &  16   &  67 &   29  & 265 & 42 & 596 &  55    & 1059 \\
		4     &  3  &  17   &  73 &   30  & 277 & 43 & 613 &  56    & 1129 \\
		5     &  5  &  18   &  79 &\bf{31}& 289 & 44 & 667 &  57    & 1153 \\
		6     &  7  &  19   &  99 &   32  & 326 & 45 & 685 &  58    & 1226 \\ 
		\bf{7}&  9  &  20   & 106 &   33  & 339 & 46 & 742 &  59    & 1251 \\
		8     & 11  &  21   & 113 &   34  & 379 & 47 & 761 &  60    & 1327 \\
		9     & 13  &  22   & 137 &   35  & 393 & 48 & 781 &  61    & 1353 \\ 
		10    & 22  &  23   & 145 &   36  & 407 & 49 & 841 &  62    & 1379 \\
		11    & 25  &  24   & 172 &   37  & 451 & 50 & 862 &\bf{63} & 1459 \\
		12    & 29  &  25   & 181 &   38  & 466 & 51 & 925 &\bf{127}& 6499 \\
		13    & 41  &  26   & 191 &   39  & 513 & 52 & 947 &\bf{255}&28323 \\
		\bottomrule
	\end{tabular}
\end{table}

\section{Satisfied $K$ is Not Sequential} \label{seq-k}

\begin{table}[h]
	\caption{Searched satisfied list of $K$ for $n=15$ and $31$.}\label{kn-seq}
	\vskip 0.15in
	\centering
	\begin{tabular}{c|c}
		\toprule
		$n:15$ & $51, 61, 62, 63, 64, 67, 68, 69, 73, 74, 75, 76, 77, 78, 79, 80, 81, 82, 83, 85\cdots$.\\
		\midrule
		$n:31$ & 
		$ \begin{array}{l}
		289, 313, 314, 315, 316, 317, 318, 326, 327, 328, 329, 339, 340, 341, 342, 343, \\
		344, 345, 346, 347, 352, 353, 354, 355, 356, 357, 358, 359, 365, 366, 367, 368,  \\
		369, 370, 371, 372, 373, 374, 375, 376, 379, 380, 381, 382, 383, 384, 385, 386,  \\
		387, 388, 389, 393, 394, 395, 396, 397, 398, 399, 400, 401, 402, 403, 404, 405,  \\
		406, 407, 408, 409, 410, 411, 412, 413, 414, 415, 416, 417, 418, 419, 421, 422\cdots, \\
		\end{array} $ \\
		\bottomrule
	\end{tabular}
\end{table}

In this section, we emphasize that not all $K\geq K_n$ satisfy Problem \ref{pro}. We show all satisfied $K$ in Problem \ref{pro} when choose $A=15$ ($n=15$) for $ 4 $-bit quantization. 
Since, from Proposition \ref{pre-res}, we know that all $K\geq\frac{(n-1)(n-3)}{2}+1=85$ satisfy Problem \ref{pro} and from Appendix \ref{all_kn}, $K_{15}=51$, we list satisfied sequences of $K$ in Table \ref{kn-seq}. 

When choose $A=31$ ($n=31$) for $ 5 $-bit quantization, $K_{31}=289$, all $ K\geq\frac{(n-1)(n-3)}{2}+1=421$ satisfies Problem \ref{pro}, we also list satisfied sequence of $K$ in Table \ref{kn-seq}.

From Table \ref{kn-seq}, there some missed numbers from $K_n$ to our upper bound $\frac{(n-1)(n-3)}{2}+1$. Moreover, hardware support may choose $K$ as power of $2$, so we recommand using $K=64$ for $4$-bit activation quantization and $K=512$ for $5$-bit activation quantization and naively $K=2^{16}$ for $8$-bit activation quantization (actually $2^{15}$ is enough).

\section{Constraint of T}\label{cons-T}
From Problem \ref{pro} and Proposition \ref{2}, we need $T\neq0$ but we do not consider this constraint in the main body of the paper. In this section, we show the case when $T=0$ satisfies.

Based on Eq. (\ref{p1}) when $T=0$, then 
$\lceil it-b \rceil=\lceil B/K \rceil$, $\forall i\in \{1,2,\dots,A\}$, which means there exists $N_0\in \mathbb{N}$,
$N_0 < t-b < At-b \leq N_0 + 1$ if we assume $t>0$.
Therefore $(A-1)t<1$, that is $0<t<\frac{1}{A-1}$.
Additionally, $\lfloor \frac{N+b}{t} \rfloor < 1 $ when $N\leq N_0$; $\lfloor \frac{N+b}{t} \rfloor\geq A $ when $ N \geq N_0+1$, showing that 
\[ \mathrm{clip}\left(\left\lfloor \frac{N + b}{t} \right\rfloor, 0, A\right) = 
0 \; or \; A. \]
We propose to use a sign function to quantize in this case.
\[ Q_{BN}(N,b,t,A)\triangleq \mathrm{clip}\left(\left\lfloor \frac{N + b}{t} \right\rfloor, 0, A\right) = A \; sign(N-N_0) =\begin{cases}
A & x > N_0 \\
0 & x \leq N_0 \\
\end{cases}, \]
where $N_0=\lceil t-b \rceil-1$ and we need to check $\lceil t-b \rceil = \lceil At-b \rceil  $.

Since the potential scope of $|t| < \frac{1}{A-1}$ is particularly limited in $\R$ when $A$ is large, hence we seldom come across this situation, and we also can solve this problem by replacing original formulation to simple binary function $Q_{BN}$ even though this case happens.

\section{Auxiliary Algorithms}\label{algo}

\begin{algorithm}[H]
	\caption{\textbf{Generate}: ($\mathcal{S}_n$) Recursively generate all possible sequences from length $n$ to length $n+1$.} \label{rec-s}
	\begin{algorithmic}[1]
		\STATE {\bfseries Input:} $\mathcal{S}_n$ which includes all possible sequences with length $n$;
		\STATE Set empty list $\mathcal{S}_{n+1}$;
		\FOR{$\{S_i\}_{i=1}^{n}$ in $ \mathcal{S}_n $}
		\STATE Generate new $ \{S_i\}_{i=1}^{n+1} $ with length $n+1$ using Proposition \ref{pro-seq} (e.g. choose $S_{n+1}=S_{n}$ or $S_{n}+1$);
		\IF{ $\max_{j>i} \frac{j(S_i-1)-iS_j}{j-i} < \min_{j<i}\frac{j(S_i-1)-iS_j}{j-i} $ }
		\STATE Add $ \{S_i\}_{i=1}^{n+1} $ to $\mathcal{S}_{n+1}$;
		\ENDIF
		\ENDFOR
		\STATE {\bfseries Output:} $\mathcal{S}_{n+1}$
	\end{algorithmic}
\end{algorithm}
\vskip -0.1in
\begin{algorithm}[H]
	\caption{\textbf{Find}($\{S_i\}_{i=1}^{n}$, $ K $):  Check $K$ is suitable for the given possible sequence $\{S_i\}_{i=1}^{n}$}
	\begin{algorithmic}[1]
		\STATE {\bfseries Input:} A possible sequence $\{S_i\}_{i=1}^{n}$ from $\mathcal{S}_n$, $K\in \mathbb{N}$;
		\FOR{\ $ T \in \mathbb{N}$ and  $\frac{KS_n-2K+1}{n-1} \le T \le \frac{KS_n-1}{n-1} $}
		\IF{$\max_{i} \left( iT- KS_i \right) < \min_{i} \left( iT- KS_i\right) + K $}
		\STATE $B = \max_{i} \left( iT- KS_i \right) $;
		\STATE {\bfseries Output:} $T, B$
		\ENDIF
		\ENDFOR
		\STATE {\bfseries Output:} $ None $
	\end{algorithmic}
\end{algorithm}
\vskip -0.1in
\begin{algorithm}[H]
	\caption{\textbf{Search}: ($\mathcal{S}_n^o$,  $ K_0 $): Search $K$ for given possible sequences in $\mathcal{S}_n^o$, a subset of $\mathcal{S}_n$}\label{search}
	\begin{algorithmic}[1]
		\STATE {\bfseries Input:} $\mathcal{S}_n^o$, a subset of $\mathcal{S}_n$, start $K_0\in \mathbb{N}$;
		\STATE $ find = False$; $ K = K_0 $;
		\WHILE{$ not $ $ find $}
		\STATE Set empty list $K_{lst}$ ;
		\FOR{$\{S_i\}_{i=1}^{n}$ in $\mathcal{S}_n^o$}
		\WHILE{\textbf{Find}($\{S_i\}_{i=1}^{n}$, $ K $) return None}
		\STATE $ K = K + 1 $;
		\ENDWHILE
		\STATE Append $K$ to $K_{lst}$;
		\ENDFOR
		\STATE Get the maximum value $K_{\max}$ from $K_{lst}$ ;
		\IF{$K_{\max} >  K $}
		\STATE $ K $ = $K_{\max}$;
		\ELSE
		\STATE $ find $ = $True$;
		\ENDIF
		\ENDWHILE
		\STATE {\bfseries Output:} $ K $
	\end{algorithmic}
\end{algorithm}

\begin{algorithm}[t]
	\caption{\textbf{Search-All} ($ \mathcal{S}_n $, $ K_0 $, $ w $): Search $K_n$ based on all possible sequences with length $ n $ in $ \mathcal{S}_n$.} \label{main-alg}
	\begin{algorithmic}[1]
		\STATE {\bfseries Input:} All possible sequences $ \mathcal{S}_n $ of sequence length $ n $ generated and checked from \textbf{Generate} function (Algorithm \ref{rec-s} in Appendix), \textbf{Search} function (Algorithm \ref{search} in Appendix), search starting point $K_0\in\mathbb{N}$ and running window $w\in \mathbb{N}$.
		\STATE $ find = False$, $ K = K_0$;
		\WHILE{$not$ $find$}
		\STATE $ s = 0 $, $ find $ = $True$;
		\WHILE{$ s < $ the length of $ \mathcal{S}_n $}
		\STATE Get $ \mathcal{S}_n^o $, a subset of $ \mathcal{S}_n$ from item index $s$ to $s+w$.
		\STATE $K_{max} = $ \textbf{Search}($ \mathcal{S}_n^o$, $ K $);
		\IF{$ K_{max} > K $}
		\STATE $ K = K_{max} $, $ find $ = $False$;
		\ENDIF
		\STATE $ s = s + w $;
		\ENDWHILE
		\ENDWHILE
		\STATE {\bfseries Output:} $ K $
	\end{algorithmic}
\end{algorithm}

\begin{algorithm}[H]
	\caption{\textbf{Get}($t, b, K, n$): Get $T,B$ in Problem \ref{pro2} when given $ t, b, K $}
	\label{getTB}
	\begin{algorithmic}[1]
		\STATE {\bfseries Input:} $t, b, n$ and $K$ from previous search.
		\STATE $ S_i = \lceil it-b \rceil, i \in \{1, 2 \cdots, n\}$.
		\STATE $T_{lower} = \max_{i>j} \Big\lfloor \frac{S_i-S_j-1}{i-j}K \Big\rfloor $.
		\STATE $T_{upper} = \min_{i>j} \Big\lceil \frac{S_i-S_j+1}{i-j}K \Big\rceil $.
		\FOR{$ T \in \mathbb{N}$ and $T_{lower} < T < T_{upper}$ (searching according to the distance to $Kt$ is faster) } 
		\STATE $B_{lower} = \max_{i} \left( iT- KS_i \right) $
		\STATE $B_{upper} = \min_{i} \left( iT- KS_i\right) + K $
		\IF{$B_{lower} < B_{upper}$}
		\STATE $B = B_{lower}$.
		\STATE {\bfseries Output:} $T, B$
		\ENDIF
		\ENDFOR
		\STATE {\bfseries Output:} $T, B$ \ (If we need to obtain all satisfied $T, B$ in Problem \ref{pro2}, we should not return when got a pair of $T, B$, but run until all possible values searched.)
	\end{algorithmic}
\end{algorithm}

\section{Other Cases}
\subsection{General Quantizer Range}\label{gene-qua}
We have already mentioned in Section \ref{pro-for} when $y_{\min}, y_{\max}$ use other integers.
Since previous discussion only focus on non-symmetric quantizer with $y_{\min}=0$ in $Q_a(\cdot)$ and $x_{\min}=0$ in $Q(\cdot)$, we briefly show ways for obtaining symmetric quantizer and more general cases in this section.

Using similar analysis in Eq. (\ref{two-affine}), we get general formulation of Problem \ref{pro}.

\begin{problem}\label{pro-gen}
	Given $Y_{\min}, Y_{\max} \in \mathbb{N}$, the problem is to find some $ K\in \mathbb{N}$ (as small as possible), such that for all $ t\neq 0, t, b \in \mathbb{R}$, there exist some $T\neq 0, T, B \in \mathbb{Z}$, such that for all $ N \in \mathbb{Z}$, 
	\[ \mathrm{clip}\left(\left\lfloor \frac{N + b}{t} \right\rfloor, Y_{\min}, Y_{\max}\right) = \mathrm{clip}\left(\left\lfloor \frac{N {\times} K + B}{T} \right\rfloor, Y_{\min}, Y_{\max}\right).\]
\end{problem}
A simple way to solve Problem \ref{pro-gen} is shifting the range of clip function.
\[  \mathrm{clip}\left(\left\lfloor \frac{N + b-tY_{\min}}{t} \right\rfloor, 0, Y_{\max}- Y_{\min}\right) = \mathrm{clip}\left(\left\lfloor \frac{N {\times} K + B-TY_{\min}}{T} \right\rfloor, 0, Y_{\max}- Y_{\min}\right).\]
Hence we are able to apply conclusions of Problem \ref{pro} with changed $b'=b-tY_{\min}$, $B'=B-TY_{\min}$ and same $t, T$ as follows:
\[  \mathrm{clip}\left(\left\lfloor \frac{N + b'}{t} \right\rfloor, 0, Y_{\max}-Y_{\min}\right) = \mathrm{clip}\left(\left\lfloor \frac{N {\times} K + B'}{T} \right\rfloor, 0, Y_{\max}-Y_{\min} \right).\]

However, due to limited numerical precision, our searching method already introduces error. We prefer to using our derivation to Problem \ref{pro-gen} directly.  

Accordingly, we obtain a simple version of Problem \ref{pro-gen}.
\begin{prop}\label{123}
	Problem \ref{pro-gen} is equivalent to finding such a $K \in \mathbb{N}$ that for all $ t \neq 0, t, b \in \R $, we can obtain $T \neq 0, T, B\in \mathbb{Z}$, s.t.
	\begin{equation}
	\lceil it-b \rceil = \Big\lceil \frac{iT-B}{K} \Big\rceil,\ \forall i \in \{Y_{\min}+1, \dots, Y_{\max}\}.
	\end{equation}
\end{prop} 
Because we know Problem \ref{pro-gen} and Problem \ref{pro} are same from previous analysis, meaning that the searched $K_n$ with $n={Y_{\max}-Y_{\min}}$ already satisfies Problem \ref{pro-gen}.
Based on Proposition \ref{tbk}, to establish $T, B$ given $K$ that is suitable based on the previous understanding, we also offer a conservative way to obtain all appropriate $T, B$ as follows.
\begin{prop}\label{tbk-gen}
	Given $t, b \in \R$ and proper $K\in\mathbb{N}$, a pair of $T, B \in \mathbb{N}$ satisfy Problem \ref{pro-gen} if and only if there exists such a $T$  that the following conditions met:
	\begin{equation}
	\begin{aligned}
	\max_{i>j}\Big\lfloor\frac{S_i-S_j-1}{i-j}K \Big\rfloor &< T < \min_{i>j} \Big\lceil \frac{S_i-S_j+1}{i-j}K \Big\rceil, \\
	\max_{i} \left( iT- KS_i \right) &< \min_{j} \left( jT- KS_j\right) + K,
	\end{aligned}
	\end{equation}
	where $i, j \in \{Y_{\min}+1, \dots, Y_{\max}\}$.
\end{prop}

In other words, our proposed method and results can simply convert to general quantizer range.

\subsection{Special Input Range}\label{sp-case}
Before training, we have little understanding of final BN parameters and hidden layer outputs, hence we consider all possible value which $b, t, N$ can take. Actually, due to quantized activation and quantized weights function in previous layers, the input $N$ varies in a finite range, but we are hard to extract effective information because $t, b$ and $N$ are combined together, so we still use whole integers for $N$.

Besides, some structures may lead to special input range of $N$ which would reduce the magnitude of $K$. For example, $N$ is always even, then the problem changes into
\[ \mathrm{clip}\left(\left\lfloor \frac{2N + b}{t} \right\rfloor, Y_{\min}, Y_{\max}\right) = \mathrm{clip}\left(\left\lfloor \frac{2N {\times} K + B}{T} \right\rfloor, Y_{\min}, Y_{\max}\right), \forall N \in \mathbb{N}.\]
Under slight transformation,
\[ \mathrm{clip}\left(\left\lfloor \frac{N + b/2}{t/2} \right\rfloor, Y_{\min}, Y_{\max}\right) = \mathrm{clip}\left(\left\lfloor \frac{N {\times} (2K) + B}{T} \right\rfloor, Y_{\min}, Y_{\max}\right), \forall N \in \mathbb{N}.\]
Hence we can use our searching method for $t/2$ and $b/2$ first, with $2K$ as satisfied solution in Problem \ref{pro-gen}. Finally, we still can obtain $T, B$ but with less $K$ used. For instance, during $4$-bit quantization, $K_{15}=51$, we may use $K=64$ for friendly hardware support. When $N$ is always even, we can use $K=32$ instead.

Generally, when input range is described as $\alpha N+\beta$ with fixed $\alpha, \beta \in \mathbb{N}$ and various $N \in \mathbb{N}$, we are able to screen out $K$ if $\alpha K$ already satisfies Problem \ref{pro-gen}. 
In conclusion, our method with special structure of input may introduce further less $K$ we searched.

\section{Quantization for CIFAR-10 Classification}\label{app-cifar}

We compare our method using VGG11 \cite{simonyan2014very} with BN on CIFAR-10 dataset.
We adopt data augmentation including horizontal flip, random cropping, random rotation and random scaling with Adam \cite{kingma2014adam} optimizer using circular learning initialized from $0.003$ for total $640$ training epochs and batch size 256. The main results are shown in Table \ref{ex_cifar}, which agree with the results on ImageNet.

\begin{table}[H]
	\caption{Top-1 test accuracy of different quantized weight and activation bits with our transformation on CIFAR-10 trained by VGG11.} \label{ex_cifar}
	\vskip 0.15in
	\begin{center}
		\begin{tabular}{c|c|c|c}
			\toprule
			Bits     &  2W4A &  4W4A &  8W8A  \\
			\midrule
			$ QT $   & 90.60 & 91.75 & 91.74  \\
			$ bt $   & 90.40 & 91.71 & 91.67  \\
			$ BT$    & 90.40 & 91.71 & 91.67  \\
			\midrule	
			\multicolumn{4}{c}{VGG11}\\
			\bottomrule
		\end{tabular}
	\end{center}
\end{table}
\end{document}